\documentclass[10pt,twocolumn,letterpaper]{article}

\usepackage{iccv}
\usepackage{times}
\usepackage{epsfig}
\usepackage{graphicx}
\usepackage{amsmath}
\usepackage{amssymb}
\usepackage[boxruled,linesnumbered]{algorithm2e}
\usepackage{comment}
\usepackage{bbm}
\usepackage{xcolor}
\usepackage{graphicx}
\graphicspath{ {./figures/} }
\usepackage{subcaption}
\usepackage{enumitem}
\usepackage[title,toc,titletoc,page]{appendix}

\usepackage[pagebackref=true,breaklinks=true,letterpaper=true,colorlinks,bookmarks=false]{hyperref}

\newcommand{\rand}{\textsc{RAND}}
\newcommand{\isotonic}{\textsc{ISO}}
\newcommand{\platt}{\textsc{PLATT}}
\newcommand{\herding}{\textsc{HERDING}}
\newcommand{\sawade}{\textsc{SAWADE}}
\newcommand{\topk}{\textsc{TOP-K}}
\newcommand{\gmm}{\textsc{GMM}}
\newcommand{\ours}{\textsc{ACIS}}
\newcommand{\oursmulti}{\textsc{ACIS}}
\newcommand{\ourslast}{\textsc{ACIS-LAST}}
\newcommand{\imagenetM}{\emph{ImageNet1M}}
\newcommand{\imagenetK}{\emph{ImageNet50K}}
\newcommand{\inat}{\emph{iNat100K}}

\iccvfinalcopy 

\def\iccvPaperID{8077} 

\ificcvfinal\fi

\def\@testdef #1#2#3{%
\def\reserved@a{#3}\expandafter \ifx \csname #1@#2\endcsname
\reserved@a  \else
\typeout{^^Jlabel #2 changed:^^J%
\meaning\reserved@a^^J%
\expandafter\meaning\csname #1@#2\endcsname^^J}%
\@tempswatrue \fi}

\begin{document}

\title{
  \vspace{-2.5em}
  Low-Shot Validation: Active Importance Sampling \\ for Estimating Classifier Performance on Rare Categories}

\vspace{-2.5em}

\author{Fait Poms*$^1$ \\
\and
Vishnu Sarukkai*$^1$\\
\and
Ravi Teja Mullapudi$^2$\\
\and
Nimit S. Sohoni$^1$\\
\and
William R. Mark$^3$\\
\and
Deva Ramanan$^{2,4}$\\
\and
Kayvon Fatahalian$^1$\\
}
\date{\vspace{-5ex}}
    \maketitle 
\newcommand\blfootnote[1]{%
  \begingroup
  \renewcommand\thefootnote{}\footnote{#1}%
  \addtocounter{footnote}{-1}%
  \endgroup
}
\blfootnote{* Both authors contributed equally to this paper}
\blfootnote{$^1$ Stanford University}
\blfootnote{$^2$ Carnegie Mellon University}
\blfootnote{$^3$ Google}
\blfootnote{$^4$ Argo AI}

\maketitle
\ificcvfinal\thispagestyle{empty}\fi

\vspace{-2.5em}
\begin{abstract}
  \vspace{-1.0em}
For machine learning models trained with limited labeled training data, validation stands to become the main bottleneck to reducing overall annotation costs.
We propose a statistical validation algorithm that accurately estimates the F-score of binary classifiers for rare categories, where finding relevant examples to evaluate on is particularly challenging.
Our key insight is that simultaneous calibration and importance sampling enables accurate estimates even in the low-sample regime ($<300$ samples).
Critically, we also derive an accurate single-trial estimator of the variance of our method and demonstrate that this estimator is empirically accurate at low sample counts, enabling a practitioner to know how well they can trust a given low-sample estimate.
When validating state-of-the-art semi-supervised models on ImageNet and iNaturalist2017, our method achieves the same estimates of model performance with up to 10$\times$ fewer labels than competing approaches. 
In particular, we can estimate model F1 scores with a variance of $0.005$ using as few as $100$ labels. 

\end{abstract}

\vspace{-2em}
\section{Introduction}

As model training techniques become increasingly label efficient,
model validation stands to become a dominant fraction of overall data annotation costs.
For example, state-of-the-art 
semi-supervised~\cite{Chen2020SimCLRv2,Jean2020BYOL,Caron2020SwAV}, 
weakly supervised~\cite{Ratner2017snorkel}, 
few-shot~\cite{Finn2017model,Nichol2018first}, 
and active learning~\cite{Settles2012active,Cohn1996active,Gal2017deep,Sener2017active} techniques 
all offer the promise of training models using a small number of human-labeled examples,
but validating the resulting models typically uses large, human-annotated datasets.
As a result, the cost of annotating validation sets is a significant factor limiting rapid model development.


In this paper we focus on the challenge of efficiently validating \emph{binary image classifiers for rare categories} (positive instances are $\leq 0.1\%$ of the dataset). 
Building binary classification models for rare categories is common in real-world settings---wildlife preservation monitoring requires identifying rare flora and fauna species; autonomous vehicles must recognize rare categories, like baby strollers, to avoid collisions. 
The validation problem is particularly difficult for rare categories:
while it is easy to collect a large amount of unlabeled data, finding even a small number of positives via uniform random sampling can require labeling thousands of images.


\begin{figure}
  \includegraphics[width=\linewidth]{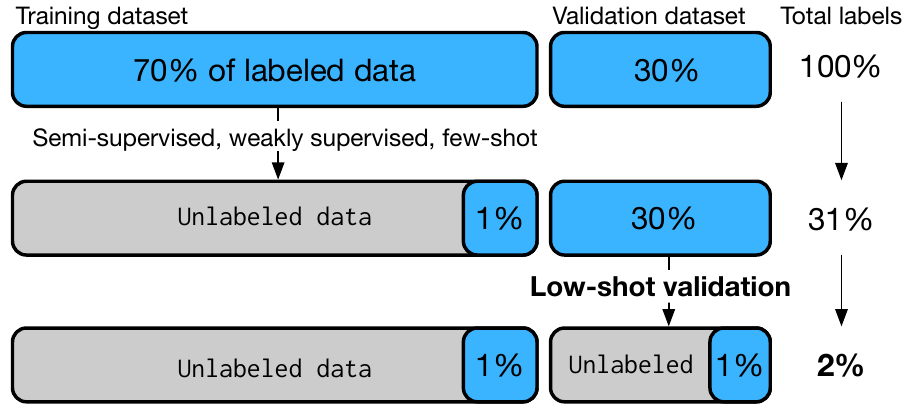}
  \vspace{-1.5em}
  \caption{Recent model training techniques such as self-supervised learning, few-shot learning, and weakly supervised learning have made it possible to train models with a fraction of the traditional fully supervised training set. However, these methods still mostly evaluate using a large validation set. In this paper, we focus on \emph{low-shot validation}, which addresses the high relative cost of collecting labeled validation data for models trained using label-efficient techniques.}
  \label{fig:teaser}
  \vspace{-2em}
\end{figure}

{\em Given a binary classification model to validate 
and a large unlabeled dataset, our goal is to estimate 
the model's F-score~\cite{FScore} on the data using a small number of annotated data samples.}
The F-score of a model depends on the distribution of the model's predicted labels, which are known, and the dataset's ground-truth labels, which require data annotation.
\emph{Importance sampling}~\cite{Tokdar2010importance} is a powerful theoretical tool for stochastically sampling the most important points in a dataset to label, but the efficiency of estimating F-scores using importance sampling depends on accurate knowledge of the likelihood that a given sample is a positive.
%
Therefore, the key challenge in using importance sampling for efficiently computing F-scores is \emph{model calibration},
the task of predicting the likelihood that a given sample is a positive, conditioned on the model scores.
Given this observation, we propose an active sampling algorithm that alternates between acquiring labels used to
train an isotonic regression model~\cite{Bianca2002Isotonic} for calibrating model probabilities, then using the calibrated model scores to 
importance-sample batches of data for metric estimation.  Using this alternating strategy, our scheme generates progressively better estimates of the model's F-score. 

We demonstrate that, particularly in the low-sample ($<$~$300$ labeled samples) regime, our algorithm can estimate F1 with significantly lower error than a variety of baselines, including semi-supervised Gaussian Mixture Models~\cite{Miller2018GMM}, prior importance-sampling approaches~\cite{Sawade2010ActiveFMeasures}, and ``herding" algorithms~\cite{Welling2009herding}. 
Not only are we able to estimate F1 efficiently, we are also able to estimate the \emph{variance} of our estimate accurately, even in low-sample regimes.
This contribution has important practical ramifications in that it allows a practitioner to know if they should trust the estimate generated by a small set of labeled validation data. Our contributions are as follows:

\vspace{-1.0em}
\begin{enumerate}[leftmargin=*,labelsep=0.5em,itemsep=-0.25em]
  \item An algorithm for joint active calibration and importance sampling-based F-score estimation. Our algorithm produces accurate and reliable estimates of a model's F-score, and it significantly outperforms baseline methods in low-sample regimes ($<300$ labeled samples).
  \item A single-trial estimator of variance for our method. We demonstrate that our variance estimator is empirically accurate, even for low-sample counts, offering a valuable diagnostic tool when using our algorithm in real-world settings. 
  \item A study that demonstrates that validation sets chosen specifically for a given model can also efficiently validate \emph{other models} trained for the same task. 
\end{enumerate}
\vspace{-1.0em}

\section{Related Work}
Approaches to label-efficient validation include statistical importance sampling-based methods~\cite{Sawade2010ActiveFMeasures}, indirect techniques that estimate precision-recall curves~\cite{Miller2018GMM,Welinder2013Semisupervised}, active learning adapted for validation~\cite{Rahman2020efficient}, and stratified sampling techniques~\cite{Bennet2010Online,Li2019Boosting,Yu2019Optimal}.
While many methods attempt to solve the validation problem, very few do so for highly imbalanced rare categories with a very small labeling budget.
We compare against a representative subset of methods which tackle this problem in our evaluation section (Sec.~\ref{sec:eval}). We delineate these methods below. 

\paragraph{Importance sampling.}
Importance sampling allows for Monte Carlo estimation of metrics using samples drawn from arbitrary distributions. Sawade et al.~\cite{Sawade2010ActiveFMeasures} propose an importance-sampling algorithm to actively estimate F-measures~\cite{FScore}, deriving an importance distribution based on the model's predicted probabilities and labels. Their method is statistically consistent but relies on assumptions of good model calibration. Our importance-sampling distribution is based on Sawade et al., and we compare against theirs in our evaluation.
\vspace{-1.5em}
\paragraph{Estimating precision-recall curves.}
Instead of estimating F-score directly, learning the shape of the precision-recall curve can help calculate a variety of validation metrics indirectly. Miller et al.~\cite{Miller2018GMM} fit the distributions of positive and negative samples across the score distribution with Gaussian mixture models (\textbf{GMM}), while Welinder et al.~\cite{Welinder2013Semisupervised} train a generative Bayesian model on the classifier's confidence scores. 
We evaluate against the \textbf{GMM} method of Miller et al.\ in order to compare against this class of techniques.
Other methods for estimating PR curves make strong assumptions which only apply once hundreds of samples have already been labeled~\cite{Sabharwal2017HowGood}. It is interesting future work to combine our low-sample-count validation with such methods which focus on validation with a budget of thousands of samples.
\vspace{-1.5em}
\paragraph{Covering the data distribution.}
``Herding" attempts to reconstruct the sufficient statistics of the dataset from a set of pseudo-random samples~\cite{Welling2009herding}. These samples can then be used to estimate F-score. We evalute against Herding in our evaluation.
\vspace{-1.5em}
\paragraph{Active learning and stratified sampling for validation.}
Active learning techniques use a partially-trained model to select samples to label that will maximize the trained performance of that model, and have recently been applied to validation~\cite{Rahman2020efficient}.
Stratified sampling techniques produce low-variance estimates by subdividing the domain into ``strata" of samples that are similar to one another, sampling from each stratum to ensure that the domain is covered~\cite{Bennet2010Online, Li2019Boosting, Yu2019Optimal}.
However, none of these techniques focus on rare categories and either apply only after an initial seed set of hundreds to thousands of labels have been collected or assume labeling budgets in the thousands.


\vspace{-0.5em}
\subsection{Other validation settings.}
\vspace{-0.5em}
Other validation settings are distinct from the one we study here, but share the challenge of label-efficient validation.
\textbf{Validation under domain shift:} When faced with domain shift in production settings, Taskazan et al.~\cite{Taskazan2020Grandfathers} show that measuring shifts in production data distributions and training models to measure the uncertainty in model prediction both help guide label-efficient and accurate validation. \textbf{Validation under noisy annotation:} Nguyen et al.~\cite{Nguyen2018ActiveTesting} explore the problem of estimating AP from noisy labels, and they introduce an active algorithm for choosing samples to label. 
\vspace{-0.5em}


\subsection{Calibration.}
\vspace{-0.5em}
With a perfectly calibrated model, it is possible to statistically estimate the number of true positives, false positives, etc. in the dataset, and thus the F-score.
Guo et al.~\cite{Guo2017NNCalibration} illustrate that larger and better-performing neural networks are often poorly-calibrated.
Platt scaling~\cite{Platt1999probabilistic} and isotonic regression~\cite{Bianca2002Isotonic} provide two well-studied~\cite{Niculescu2005predicting} methods of calibrating models, and we compare against both techniques in our evaluation.
We utilize isotonic regression for calibration in our iterative algorithm. 

\vspace{-0.5em}
\subsection{Classifier training with few manual labels.}
\vspace{-0.5em}
Recent advances in semi-supervised training can produce models trained with just 1\% of the ImageNet training set labels that are competitive with models trained on the fully supervised ImageNet dataset~\cite{Caron2020SwAV, Henaff2020Contrastive, Jean2020BYOL, falcon2020framework, Chen2020SimCLRv2}.
Semi-supervised methods produce models using a blend of manual and \emph{automatic} labels~\cite{Schmarje2020Survey},
However, their performance is evaluated using the \emph{entire} ImageNet validation split (50,000 labeled images), tempering the actual reduction in overall annotation budget achieved. Our goal is label-efficient validation of these methods, and we validate three state-of-the-art semi-supervised models in Section~\ref{sec:eval}.

\begin{figure}[t]
  \includegraphics[width=\linewidth]{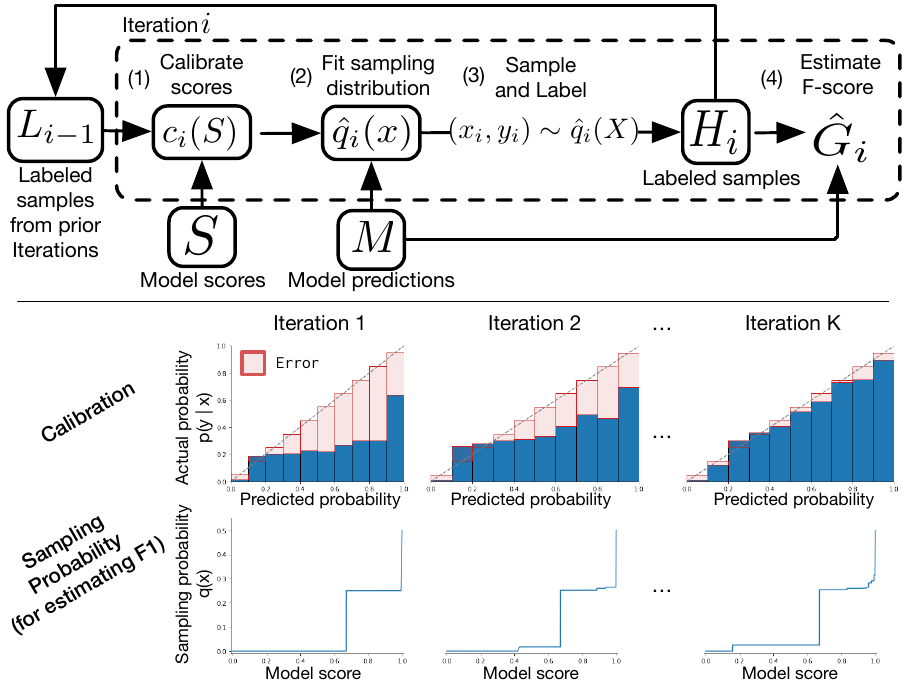}
  \vspace{-1.5em}
  \caption{We estimate a model's F-score with a small labeling budget using an iterative algorithm which combines classifer calibration and importance sampling. \textbf{Top:} each iteration of our algorithm (1) calibrates the model's scores using the labels from the prior iteration, (2) uses these calibrated scores and the models' predictions to fit an importance sampling distribution, (3) samples a batch from this distribution to label, and (4) uses these samples to estimate the model's F-score. \textbf{Bottom:} visualizations of the intermediate iterations of our method on step (1) calibration and step (2) the importance sampling distribution. As our labeling budget increases, the model's scores gradually converge to a calibrated model (linear gray dotted line) and the sampling distribution becomes increasingly refined. The sampling distribution is optimized to compute the model's F1. } 
  \label{fig:method_overview}
  \vspace{-1.5em}
\end{figure}

\vspace{-0.5em}
\section{Method}
\vspace{-0.5em}
Our goal is to estimate the F-score~\cite{FScore} for a target model on a (potentially infinite) test distribution given a small labeling budget.
Figure~\ref{fig:method_overview} provides an overview of our method: given the model's predictions and scores, we iteratively improve our estimate of the model's F-score by repeatedly (1) calibrating the model's raw scores $S$ to produce better estimates $c(s)$ of the probability that a sample is a positive, (2) using the calibrated scores to compute an importance sampling distribution $\hat{q}(x)$ that prioritizes samples most important for accurately estimating the F-score, (3) labeling samples drawn according to $\hat{q}(x)$ and (4) using those samples to estimate the F-score using importance sampling.
In this section, we first introduce our problem setup more formally and provide the necessary background on importance sampling for estimating F-scores. 
Then, we describe our method and how to estimate the method's variance accurately with a single trial.  

\makeatletter
\newcommand{\funclabel}[1]{%
  \@bsphack
  \protected@write\@auxout{}{%
    \string\newlabel{#1}{{\jayden@currentfunction}{\thepage}}%
  }%
  \@esphack
}
\makeatother

\vspace{-0.5em}
\subsection{Problem Statement}
\vspace{-0.5em}
Given a model $M$ and a dataset $(x_j,y_j)\sim X,Y, j\in\{1...n\}$ with features $x_j$ and unknown true labels $y_j$, our goal is to estimate $M$'s performance at predicting the true labels on $X$ given a finite labeling budget. In this paper, we focus on estimating the F-score, $G$, of the model, a measure of the deviation between labels predicted by the model $\hat{Y}$ and the true labels $Y$. Parameterized by $\alpha \in [0,1]$, The F$_\alpha$-score $G$ \cite{FScore} is defined:
\vspace{-0.5em}
\begin{equation}
  G = \frac{tp}{\alpha(tp + fp) + (1 - \alpha)(tp + fn)}
\end{equation}
where $tp = E[\mathbbm{1}[Y = 1 \land \hat{Y} = 1]]$ is the model's true positive rate, $fp = E[\mathbbm{1}[Y = 0 \land \hat{Y} = 1]]$ the false positive rate, and $fn = E[\mathbbm{1}[Y = 1 \land \hat{Y} = 0]]$ the false negative rate. We obtain F1 by setting $\alpha=0.5$, and obtain precision and recall by setting $\alpha=1$ and $\alpha=0$ respectively.

We aim to understand the deviation between our predicted F-score $\hat{G}$ and the true F-score $G$. 
We do so by measuring the mean-squared error (MSE) of our estimation method $E[(\hat{G} - G)^2]$ and by studying its bias $E[\hat{G} - G]$ and the variance $E[\hat{G}]^2 - E[\hat{G}^2]$.

We assume the model $M$ generates predicted labels $\hat{y}_j \in \{0, 1\} \sim \hat{Y}, j\in\{1...n\}$, and model scores $s_j \in S, j\in\{1...n\}$.

\vspace{-0.5em}
\subsection{Background: Importance Sampling for Consistent Estimation of F-Score}
\vspace{-0.5em}
When estimating a metric, importance sampling is a technique that non-uniformly selects samples according to a sampling distribution $q$.
The technique aims to choose a $q$ that is optimized to produce good estimates of the metric, and it corrects for the unequal probabilities of selection in order to produce accurate estimates.
Sawade et al.~\cite{Sawade2010ActiveFMeasures} introduce an importance sampling method for the consistent (asymptotically unbiased) estimation of F-measures. 
Suppose $p(x,y)$ defines the probability distribution across the population of samples $x \in X$ and labels $y \in Y$. Let $v(x,y,\hat{y}) = \alpha \hat{y} + (1 - \alpha)y$, with $\alpha \in [0,1]$, and $\ell = 1 - \ell_{0/1}$, the zero-one loss. With any distribution $q(x,y)$ where $q$ is nonzero across the domain of $p$ and $(x_1,y_1),(x_2,y_2)... \sim q(x,y)$, Sawade et al.\ approximate the population F-score $G$ as $\hat{G}_{n,q}$: 

\begin{equation}
  \hat{G}_{n,q} = \frac{\sum_{j=1}^n w(x_j,y_j,\hat{y}_j) \ell(\hat{y}_j,y_j)}{\sum_{j=1}^n w(x_j,y_j,\hat{y}_j)}
\end{equation}
\[w(x_j,y_j,\hat{y}_j) = \frac{p(x_j)}{q(x_j)} v(x_j,y_j,\hat{y}_j)\]

They prove that $\hat{G}_{n,q}$ is a consistent estimator of $G$ i.e. $\hat{G}_{n,q} \xrightarrow[n \to \infty]{} G$. Therefore, as the sample size $n$ increases, choosing the distribution $q^*$ that minimizes the variance of $\hat{G}_{n,q}$ becomes inceasingly effective at minimizing the MSE of the estimate. 

\vspace{-0.5em}
\paragraph{Optimal sampling distribution}
\vspace{-0.5em}
Sawade et al.\ derive the theoretically optimal variance-minimizing sampling distribution, $q^*$, for their estimator:

\vspace{-1em}
\begin{equation}
    q^*(x) \propto 
      \begin{cases}
        p(x)\big(p(y=1|x)(1 - G)^2 + \\
        \quad \quad \alpha^2(1 - p(y=1|x))G^2\big)^{0.5} & :\hat{y} = 1\\
        p(x)(1-\alpha)\big(p(y=1|x)*G^2\big)^{0.5} & :\hat{y} = 0
      \end{cases}
    \label{eq:optimal}
\end{equation}

However, this formula assumes knowledge of $G$, the very metric that we aim to estimate. In addition, it assumes knowledge of $p(y|x)$, which is unknown as well. 
Sawade et al.\ work around this limitation by substituting model scores $S$ for $p(y=1|x)$, which assumes that the model being evaluated is perfectly calibrated. 
However, many neural networks are not well-calibrated~\cite{Guo2017NNCalibration}, in particular when training classifiers for rare categories~\cite{wallace2012class}. 
We address these limitations in our proposed algorithm. 

\vspace{-0.5em}
\subsection{Active Calibration and Importance Sampling}
\vspace{-0.5em}
We propose \textbf{A}ctive \textbf{C}alibration and \textbf{I}mportance \textbf{S}ampling (\ours, Algorithm~\ref{alg:algorithm}), an iterative importance sampling algorithm for simultaneously estimating $p(y=1|x)$ (calibrating the model) and estimating $G$.
Each iteration $i$ first calculates $\hat{q}_i$, our current best estimate of $q^*$, using the samples that are already labeled.  
$\hat{q}_i$ is estimated by training an isotonic regression~\cite{Bianca2002Isotonic} model $c_i(s)$ on the model scores $s \in S$ to predict $p(y=1|x)$, using the samples that are already labeled to train. 
Applying equation \ref{eq:optimal}, substituting $c_i(s)$ in place of $p(y=1|x)$ and $\hat{G}_{i-1}$ in place of $G$:

\vspace{-1em}
\begin{equation}
    \hat{q}(x) \propto
      \begin{cases}
        p(x)\big(c_i(s)(1 - \hat{G}_{i-1})^2 + &\\
        \quad \quad \alpha^2(1 - c_i(s))\hat{G}_{i-1}^2\big)^{0.5} & :\hat{y} = 1\\
        p(x)(1-\alpha)\big(c_i(s)*\hat{G}_{i-1}^2\big)^{0.5} & :\hat{y} = 0
      \end{cases}       
      \label{eq:ours}
\end{equation}

\setlength{\textfloatsep}{1pt}
\begin{algorithm}[t]
    \KwData{$\hat{Y}, S, B, \alpha$}
    \KwResult{$\hat{G}_1, \hat{G}_2,...$, $H_1, H_2...$, $W_1, W_2, ...$}
    \DontPrintSemicolon
    \SetKwFunction{FSample}{Estimate}
    \SetKwFunction{FUpdateCal}{UpdateCal}
    \SetKwFunction{FComputeF}{ComputeFScore}
    \SetKwFunction{FIso}{Iso}
    \SetKwFunction{FIsoTrain}{Iso.Train}
    \SetKwFunction{FIsoEval}{Iso.Eval}
    \SetKwFunction{FUnique}{Unique}
    \SetKwFunction{FWeightedSample}{WeightedSample}
    \SetKwProg{Fn}{Function}{:}{}
    \Fn{\FSample{$\hat{P}_y$, $budget$, $\hat{G}$}}{
            $\hat{q}$ = \textbf{ImportanceDistribution}$(\hat{P}_y, \hat{Y}, \hat{G}, \alpha)$\;
            $H, W$ = \textbf{WeightedSample}$(q, budget)$\;
            $\hat{G}$ = \textbf{FScore}$(H, \hat{Y}, W)$\;
            \KwRet $H$, $W$, $\hat{G}$\;
    }
    
    $c_0$ = \textbf{IsotonicRegression}$(S, \hat{Y})$, $\hat{G}_0 = 0.5$\;
    $B_1 = 10$, $i=1$, $L_0$ = $\{\}$\;

    \While{$|L_{i-1}|$ $<$ $B$}{
        $H_i,W_i\hat{G}_i$ = \FSample{$c_{i-1}(S), B_{i}, \hat{G}_{i-1}$} \;
        $L_i$ = $L_{i-1} \cup H_i$ \;
        $c_i$ = \textbf{IsotonicRegression}$(\hat{P}_y, L_i)$ \;
        $B_{i+1} = 2*B_{i}$, $i = i+1$\;
    }
    \KwRet $\hat{G}_1, \hat{G}_2,...$, $H_1, H_2...$, $W_1, W_2, ...$\;
    \caption{ \textsc{ACIS} }
    \label{alg:algorithm}
\end{algorithm}
\setlength{\textfloatsep}{8pt}

Isotonic regression outperforms Platt scaling~\cite{Platt1999probabilistic} in the setting with extremely imbalanced classes and classification models of varying quality (see Supplemental for details). 
The isotonic model is trained by reusing samples $L$ that are initially labeled in order to estimate $\hat{G}$. 
In the first iteration, when there are no existing labeled examples, the algorithm samples from the calibration prior $c_0(s)$, which is an isotonic model trained on the predicted labels $\hat{Y}$ rather than the true labels $Y$. 
For all iterations, the range of $c_i(s)$ is linearly rescaled from $[0,1]$ to $[\epsilon, 1 - \epsilon]$ in order to ensure that $\hat{q}_i$ is nonzero for all samples. 
Next, a new batch of samples $H_i$ drawn from $\hat{q}_{i-1}$ is first labeled, then used to compute $\hat{G}_i$, the current estimate of $G$.
At the end of the process, the algorithm returns a sequence of estimates $\hat{G}_1,\hat{G}_2,...$ of $G$. Our estimates of $G$ progressively improve as our estimate of $q^*$ improves. 
\vspace{-1.5em}
\paragraph{Estimate averaging.}
By default, $\hat{G}_k$, the last iteration returned by Algorithm \ref{alg:algorithm}, can be used as the estimate of F-score. 
However, in order to utilize the earlier samples, the final $\ell$ estimates of $G$ are aggregated as follows:
\begin{equation}
  \hat{G} = \frac{\sum_{i=k-\ell+1}^k\hat{G}_i*|W_i|_1}{\sum_{i=1}^k |W_i|_1}
  \label{eq:combineIterations}
\end{equation}
By Sawade et al.~\cite{Sawade2010ActiveFMeasures}, each of the $\ell$ $\hat{G}_i$'s are consistent estimators. Since $\hat{G}$ is a weighted average of a finite number of consistent estimators, $\hat{G}$ is also consistent. 
By averaging across the $\ell$ $\hat{G}_i$'s, we aim to improve the estimate of $G$ by incorporating information from earlier iterations of sampling, not just the last iteration.

\vspace{-1.5em}
\paragraph{Reusing prior iteration samples}
In each iteration $i$, sample-efficiency is improved by making use of all previously-labeled samples $L_{i-1}$.
Since these points have already been labeled in a previous iteration, they do not add to our overall labeling budget. To account for deterministically labeling $|L|$ points in a dataset of size $|X|$, their importance weights are set to $\frac{p(x)}{\hat{q}_i} = \frac{|L|/|X|}{|L|/|L|} = \frac{|L|}{|X|}$. 

\vspace{-1.5em}
\paragraph{Weighted average of $c_i(s)$.}
In practice, the $c_i(s)$ learned in the early iterations of the model can be unstable. We compensate for this effect by calibrating using $\beta*c_0(s) + (1 - \beta)*c_i(s)$, a weighted combination of $c_i(s)$ and the calibration prior $c_0(s)$, and decreasing the weight $\beta$ linearly over the first few iterations of the algorithm.  

\vspace{-1.5em}
\paragraph{Adaptive top-K prior for rare categories.}
When estimating F-score for rare categories, the vast majority of ground-truth positives are often high in the sorted ordering of model scores $S$. 
This means that sampling the rest of the ordering largely yields negatives. 
Therefore, when estimating model F1 for rare categories the sampling domain is restricted to the $3*(i+1)*n_{pos}$ examples with the highest score $s$, where $n_{pos}$ is the number of samples labeled positive by the model in the dataset. 
There are $n_{pos} + fn$ samples in the dataset that are relevant for estimating F-score, and it is difficult to estimate $fn$ accurately, so our sampling range heuristic is dependent on $n_{pos}$.
In addition, by making the sampling range dependent on $i$, the set of potential samples that can be labeled every iteration is expanded, slowly ``relaxing" the heuristic. 
While limiting the sampling domain introduces bias by potentially undersampling false negatives, in practice the reduction in variance far outweighs the slight bias introduced, and in low-sample regimes the domain restriction significantly reduces the MSE of our algorithm.  

\vspace{-0.5em}
\subsection{Calculating Variance}
\vspace{-0.5em}
The variance of a randomized estimator is a powerful diagnostic tool for understanding its potential error.
We derive the following consistent estimator of sampling variance for our active method:

\vspace{-1.5em}
\begin{equation}
  S^2_{n,q} = \frac {C^{-1} * \sum_{j=1}^n w(x_j,y_j,f_\theta)^2 \left( \ell(f_\theta(x_j),y_j) - \hat{G}_{n,q}\right)^2} {\frac{1}{n}\left(\sum_{j=1}^n w(x_j,y_j,f_\theta)\right)^2}
\end{equation}
\[
  C = 1 - \frac{\sum_{j=1}^n w(x_j,y_j,f_\theta)^2}{(\sum_{j=1}^n w(x_j,y_j,f_\theta))^2}
\]

Our derivation leverages the Delta Method~\cite{Doob1935limiting} to obtain a consistent estimator of sampling variance, then applies a Bessel-style correction $C$ to improve performance in low-sample regimes. The derivation is included in the Supplemental. 

When applying Equation \ref{eq:combineIterations} to combine the estimates $\hat{G}_i$'s, the variance of the weighted average $\hat{G}$ is estimated by taking a weighted average of the sampling variances of each iteration. 
The implied assumption that there is no covariance between the $\hat{G}_i$'s is a reasonable assumption in practice, and yields better estimates of the variance of $\hat{G}$ than the worst-case estimator (which assumes a covariance of 1). Experimental evidence is included in the Supplemental. 

\section{Evaluation}
\label{sec:eval}

Our evaluation compares the sample efficiency of our F-score estimation algorithm to baseline approaches. 
We also provide an analysis of our method's ability to provide bounds on the estimated metric's error by predicting the variance from just a single trial. 
We refer to our method as \ours\ (\textbf{A}ctive \textbf{C}alibration and \textbf{I}mportance \textbf{S}ampling). 

\vspace{-0.5em}
\subsection{Experimental setup}
\vspace{-0.5em}
\paragraph{Datasets.}
We evaluate \ours\ on the ImageNet~\cite{Russakovsky2015imagenet} and iNaturalist~\cite{Van2018inaturalist} large-scale image classification datasets.
ImageNet is an image classification dataset with 1000 categories, 1.2 million training images, and 50,000 validation images. 
To investigate the semi-supervised setting, we follow~\cite{Chen2020SimCLRv2} by restricting training labels to the same 1\% split of the training dataset. 
We measure validation performance on two datasets: 1) the remaining 99\% of ImageNet train dataset (which we refer to as \imagenetM) and 2) the ImageNet validation dataset (\imagenetK). 
In addition, we evaluate on iNaturalist (\inat) an image classification dataset with 5089 categories, 579,000 training images, and 95,986 validation images. 
Taking inspiration from~\cite{Chen2020SimCLRv2}, we construct a 10\% split of the training dataset for semi-supervised learning.

\vspace{-1.5em}
\paragraph{Models.}
To test our algorithm on models trained with limited labeled data, we estimate the F1 of three self-supervised learning methods which provide state-of-the-art semi-supervised performance: SwAV~\cite{Caron2020SwAV}, SimCLRv2~\cite{Chen2020SimCLRv2}, and BYOL~\cite{Jean2020BYOL}.
We treat each of the 1000-way outputs of the classifier as a binary classifier through one-vs-all classification.
Results are presented in terms of average performance across these 1000 binary classification tasks.
(We train the SwAV and BYOL models ourselves as off-the-shelf weights are not provided; details in the supplemental.)

While our experiments evaluate binary classifiers constructed from multi-class classifiers, our algorithm is capable of validating any arbitrary binary classifier that produces class scores. 
Our method only requires class scores and predicted labels from a target model. 


\vspace{-1.5em}
\paragraph{Baselines.}
We evaluate our approach against: 
\topk, an approach common in information retrieval \cite{Rahman2020efficient} which draws the top K samples from the model's ranked scores; 
\gmm~\cite{Miller2018GMM}, which also labels the top K samples from the model's ranked scores, then fits a two-component Gaussian mixture model to the model's score distribution to predict labels on the unlabeled examples; 
and \herding~\cite{Welling2009herding}, which attempts to approximate the metric using samples that can reconstruct the sufficient statistics of the dataset.

\subsection{Validating semi-supervised models}

\begin{figure}[t]
    \includegraphics[width=\linewidth]{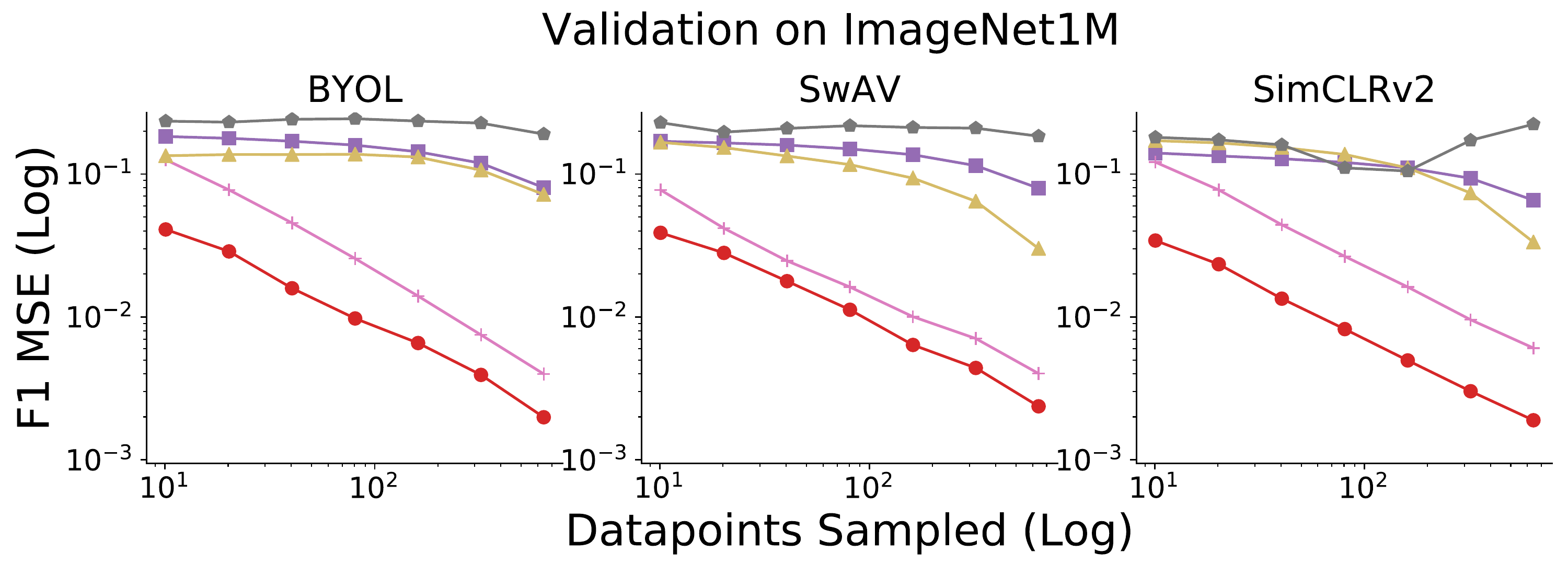}
    \includegraphics[width=\linewidth]{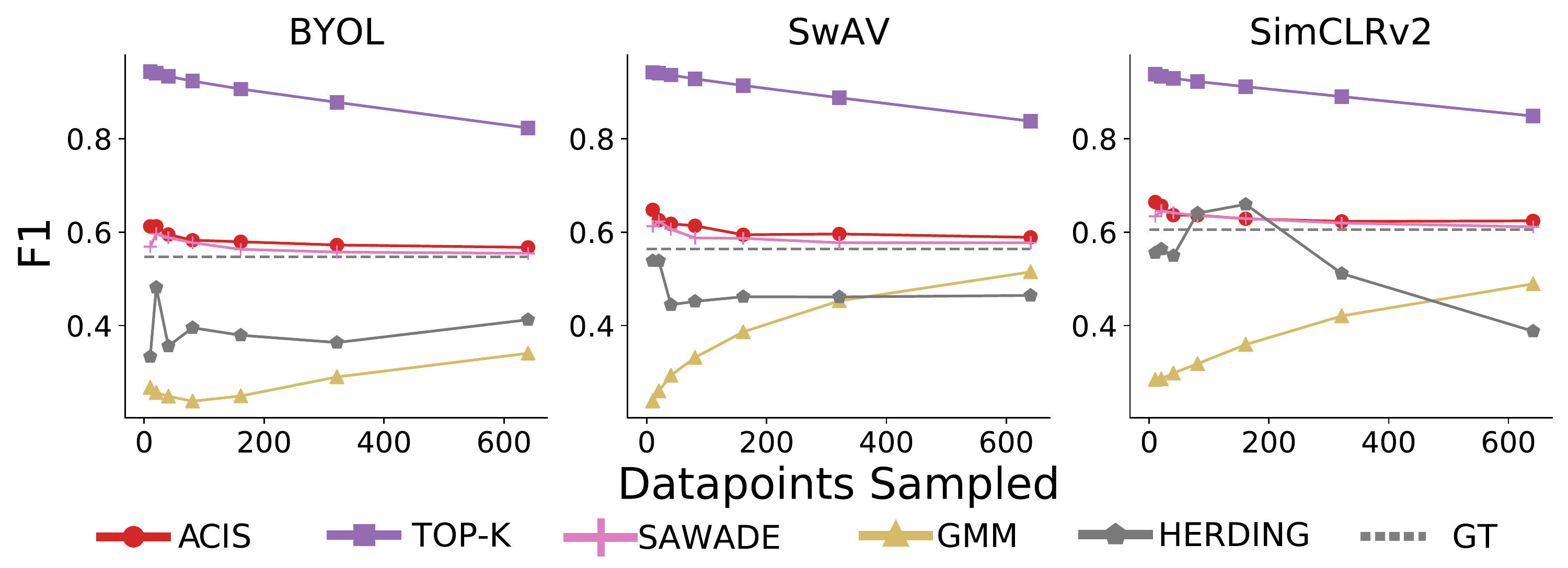}
    \vspace{-1.5em}
    \caption{
        \ours\ estimates the F1 scores of semi-supervised models to MSE $<$~$0.01$ using less than 100 samples, even when sampling from the large \imagenetM\ dataset. \textbf{Top:} the mean squared error (MSE) of the estimated F1, averaging across a single trial for each of the 1000 ImageNet categories.  \ours\ has consistently lower MSE than all baselines. \textbf{Bottom:} the predicted F1 score, averaged across a single trial for each of the 1000 ImageNet categories. In all cases, \ours\ estimates the F1 score in expectation to within 0.1 of the true value, even for as few as 10 samples. Other than \sawade, the other baselines exhibit more bias. 
    }
    \label{fig:plot3a}
    \vspace{-0.1em}
\end{figure}

\begin{figure}[t]
  \includegraphics[width=\linewidth]{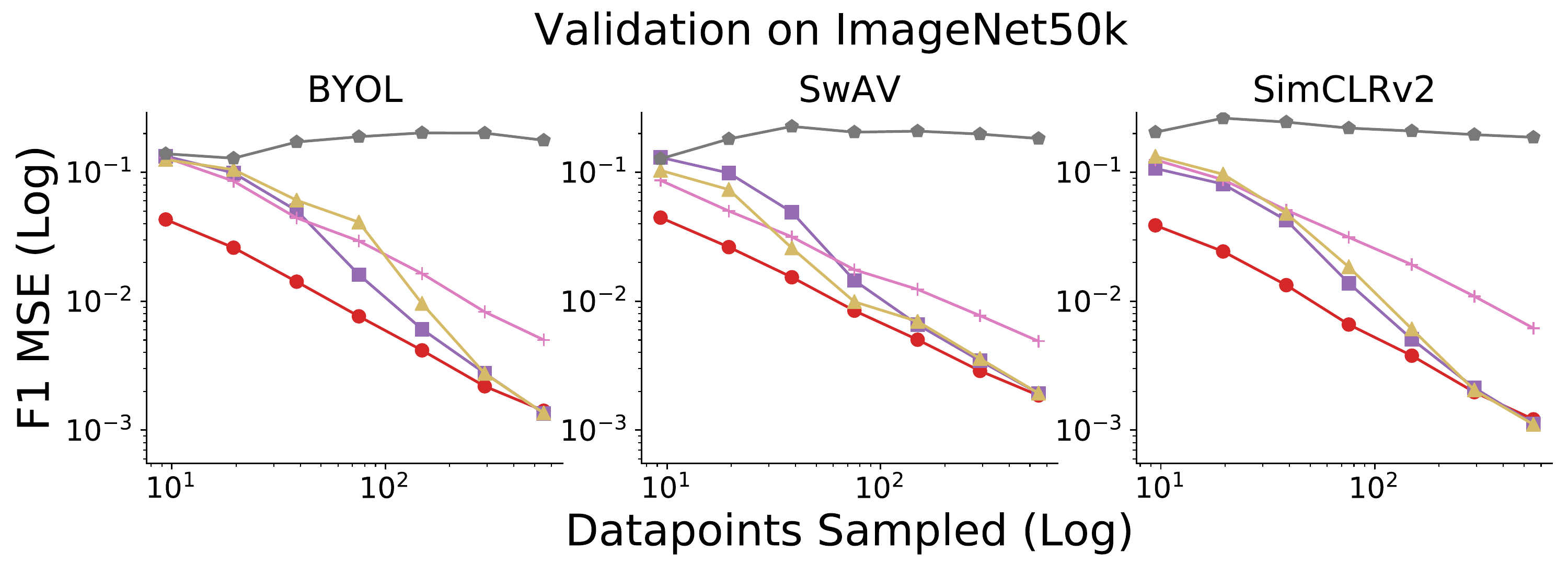}
  \includegraphics[width=\linewidth]{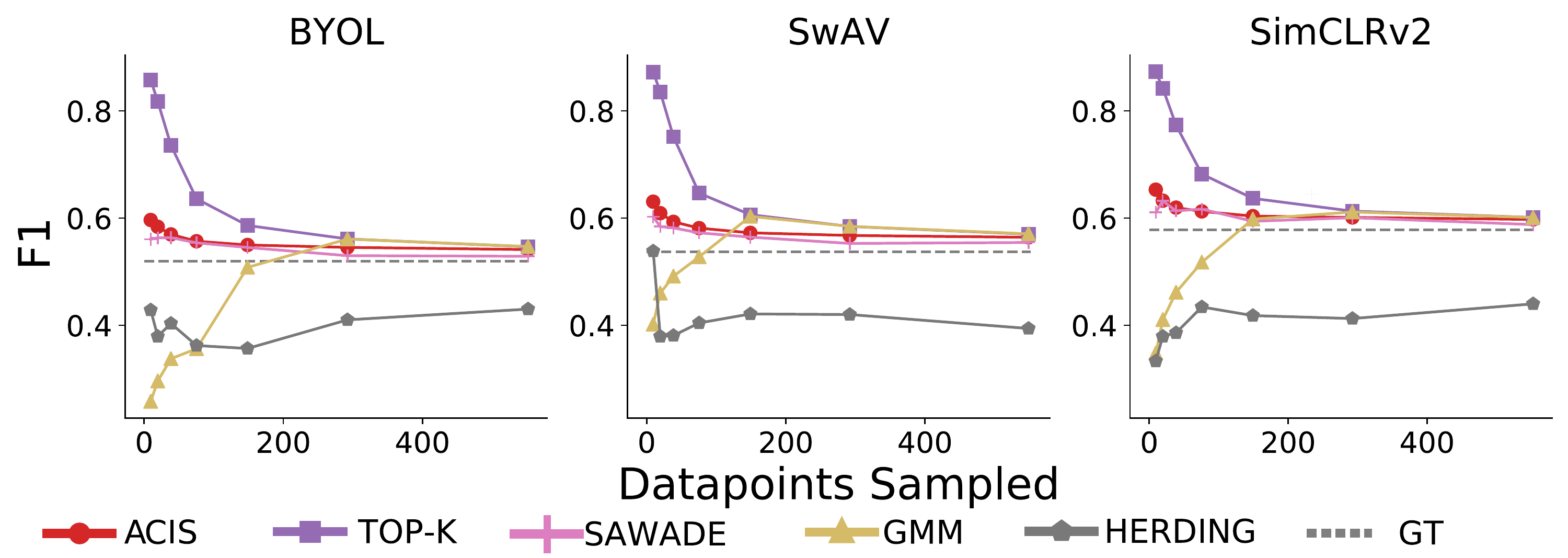}
  \vspace{-1.5em}
  \caption{
      Validation accuracy on \imagenetK\, following the experimental setup of Fig.~\ref{fig:plot3a}. Again, \ours\ generates lower MSE estimates in low-sample regimes ($<$~$300$ samples). 
      Since \imagenetK\ is 20$\times$ smaller than \imagenetM\, there are fewer relevant samples to find for a given category, so the MSE for most methods converges by $500$ samples.
  }
  \label{fig:plot3}
\end{figure}

\begin{figure}[t]
    \includegraphics[width=\linewidth]{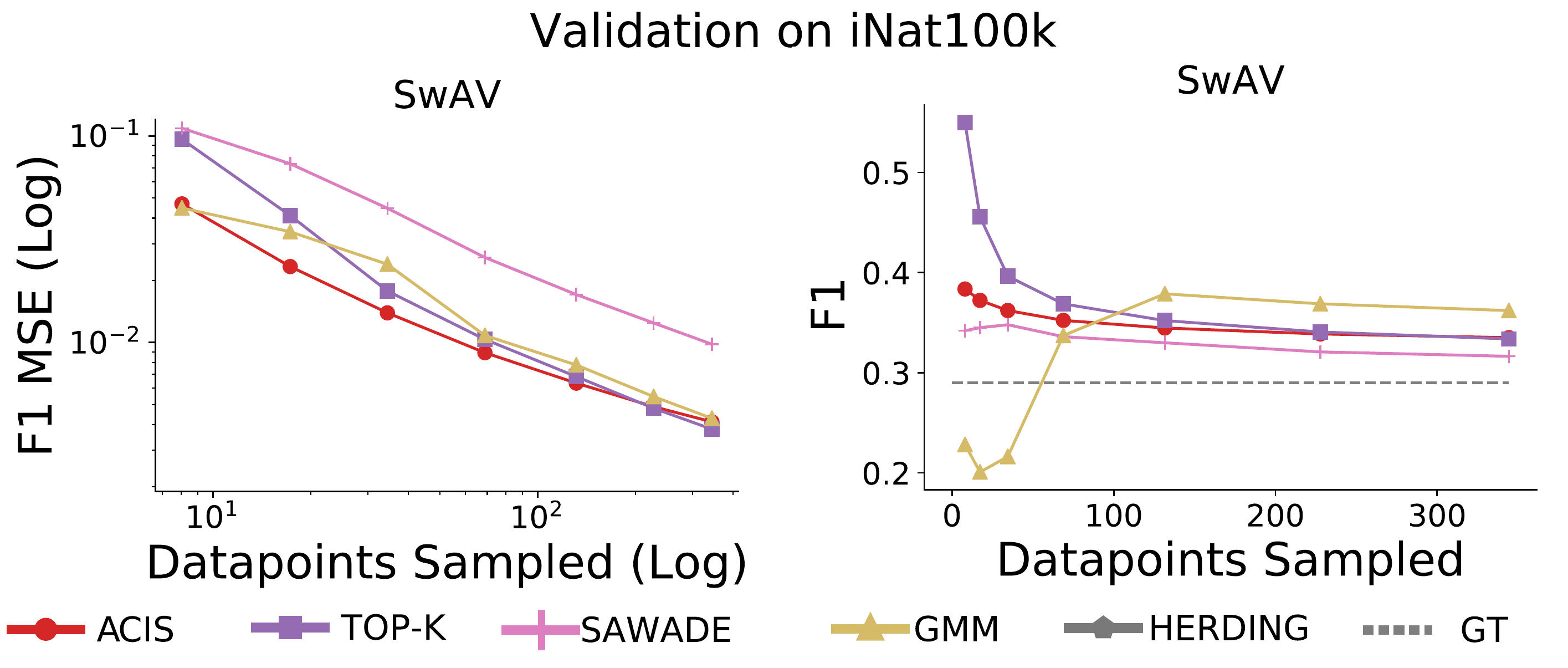}
    \vspace{-1.5em}
    \caption{
        Validation accuracy on \inat\, following the experimental setup of Fig.~\ref{fig:plot3a}. 
        In low-sample regimes ($<$~$100$ samples), F1 estimates by \ours\ on the \inat\ dataset have lower MSE than baselines. 
        \inat\ has fewer images than \imagenetM\ and fewer images per class than \imagenetK, so there are fewer relevant samples to find, so most methods converge in MSE by $200$ samples.
    }
    \label{fig:plot3inat}
    \vspace{-0.0em}
\end{figure}

\vspace{-0.5em}
\subsubsection{Comparison to baselines}
\vspace{-0.5em}

\paragraph{Validation on \imagenetM.}
We first compare the performance of the different validation methods on \imagenetM, a dataset with over a million images. Figure~\ref{fig:plot3a} gives the estimated F1 score (bottom) and mean squared error (MSE) (top) for the F1 estimate (compared to the metric computed on the full dataset) for all methods, running a single trial of each method on every ImageNet class, then averaging across the 1000 classes. 
In low-sample regimes (labeling $<$~$600$ samples), \ours\ generates estimates with significantly lower MSE than the baseline approaches. 
Estimates by \sawade\ have slightly lower bias than \ours, but \sawade\ performs worse in terms of MSE since it constructs its sampling distribution using uncalibrated probabilities. 
\topk\ is not competitive because it fails to sample false negatives when sampling a small fraction of a million-image dataset.
Similarly, \gmm\ appears to lack sufficient signal in the tail of the distribution in order to fit the mixture model for the distribution of positive and negatives samples. 
\herding\ performs poorly, perhaps because the task of generating pseudo-random samples to estimate the moments of a high-dimensional distribution does not translate well to the task of estimating F-score in low-sample regimes.

\vspace{-1.5em}
\paragraph{Validation on \imagenetK.}
We also compare the validation methods on the smaller \imagenetK\ dataset (50,000 unlabeled samples).
As seen in Figure~\ref{fig:plot3}, \ours\ consistently outperforms baseline approaches when labeling less than 300 samples for a binary classifier. 
This low-sample regime is particularly significant because the classifiers being evaluated are only trained on 10 positive samples per class.
\topk\ and \gmm\ have high bias in the low-sample regime due to consistently overestimating or underestimating the model's F1, respectively (Fig.~\ref{fig:plot3}-bottom).
As the labeling budget increases, \topk\ and \gmm\ converge to \ours\ because most of the relevant samples for the F1 score have been labeled.
\ours\ performs similarly in terms of MSE on both \imagenetK\ and \imagenetM, as seen in a comparison of Figures~\ref{fig:plot3a} and \ref{fig:plot3}. 
\vspace{-1.5em}
\paragraph{Validation on \inat.} 
We also compare the validation methods on the \inat\ dataset, validating a model trained with SwAV. 
The typical class in \inat\ has approximately 20 instances of each class, whereas \imagenetK, a dataset of 50,000 images, has $\frac{50000}{1000}$=50 images per class. 
As a result, in Figure~\ref{fig:plot3inat}, we observe trends similar to \imagenetK, but methods converge after a smaller number of samples (100, as opposed to 300 for \imagenetK).
Similar to the \imagenetK\ experiments, \topk\ and \gmm\ compare the most favorably to \ours.


\vspace{-1.0em}
\subsubsection{Ablation analysis}
\vspace{-0.5em}

We perform an ablation analysis to understand the benefits of different components of \ours.
To ablate the averaging of F1 estimates from several iterations, \ourslast\ uses the F1 estimate from only the last iteration of \ours. 
Unlike \oursmulti, \sawade~\cite{Sawade2010ActiveFMeasures} ablates model calibration.  
We also ablate importance sampling in three configurations:
\isotonic\ samples at uniform random, then applies isotonic regression~\cite{Bianca2002Isotonic} to infer labels on unlabeled data points.
\platt\ samples at uniform random, then applies Platt scaling~\cite{Platt1999probabilistic} to infer labels on unlabeled data points; and
\rand\ samples at uniform random, then estimates the metric using only the selected samples.

\begin{figure}
    \includegraphics[width=\linewidth]{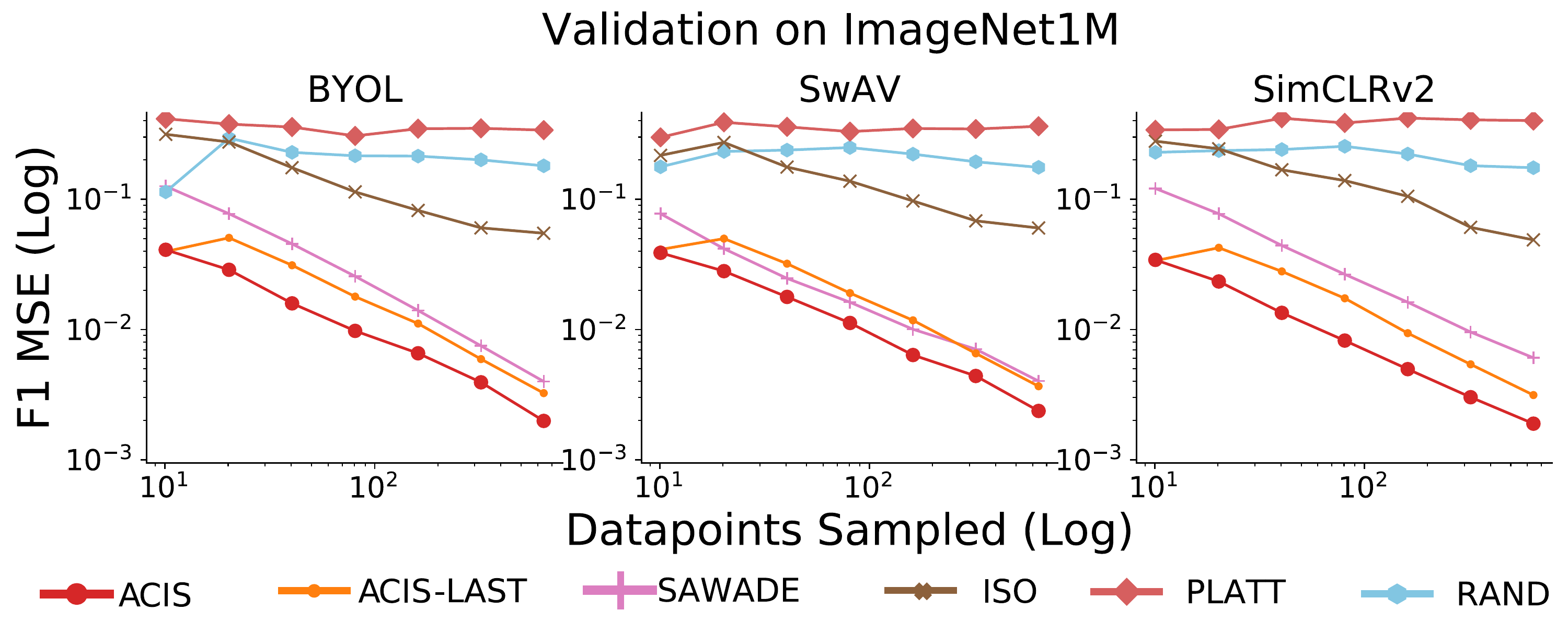}
    \vspace{-1.5em}
    \caption{Ablation analysis: \oursmulti, which jointly performs calibration and importance sampling, performs better than both calibration-only approaches (\isotonic, \platt) and importance sampling without calibration (\sawade). Taking a weighted average of our iterative estimates of the F1 also outperforms just taking the last (\oursmulti\ vs \ourslast\ respectively). All approaches except \platt\ significantly outperform uniform random sampling (\rand).}
    \label{fig:factor_analysis}
\end{figure}

Figure~\ref{fig:factor_analysis} shows the MSE of the various methods on \imagenetK.
Jointly calibrating (\oursmulti) is more sample-efficient than a pure importance sampling method (\sawade) because the importance sampling distribution improves with well-calibrated models. 
Only using model calibration (\isotonic, \platt) performs no better than uniform sampling (\rand) in the low-data regime ($<$~$100$ samples) because it does not actively select samples to improve the estimate of the metric.
Jointly calibrating and performing importance sampling to estimate F1 is more effective than using either technique in isolation.
Combining estimates from multiple iterations has a small benefit--\oursmulti\ outperforms \ourslast\ in the regime with fewer than $100$ samples, though the methods converge for larger sample sizes.

\vspace{-1.0em}
\subsubsection{Computational cost}
\vspace{-0.5em}

Compared to training, inference, and labeling costs of the model development process, the computational costs of \oursmulti\ are trivial. 
The entirety of the computational cost of iterative calibration and sampling is less than a minute per model in our experiments on a single CPU. 
This small computation cost can potentially yield a significant reduction in the number of images that must be hand-annotated to achieve a target validation accuracy.

\vspace{-0.5em}
\subsection{Variance Diagnostics for Estimating F-score}
\vspace{-0.5em}

\subsubsection{Variance Estimation for \ours}
\vspace{-0.5em}

\begin{figure}[t]
    \begin{center}
    \begin{subfigure}{.96\linewidth}
        \includegraphics[width=\linewidth]{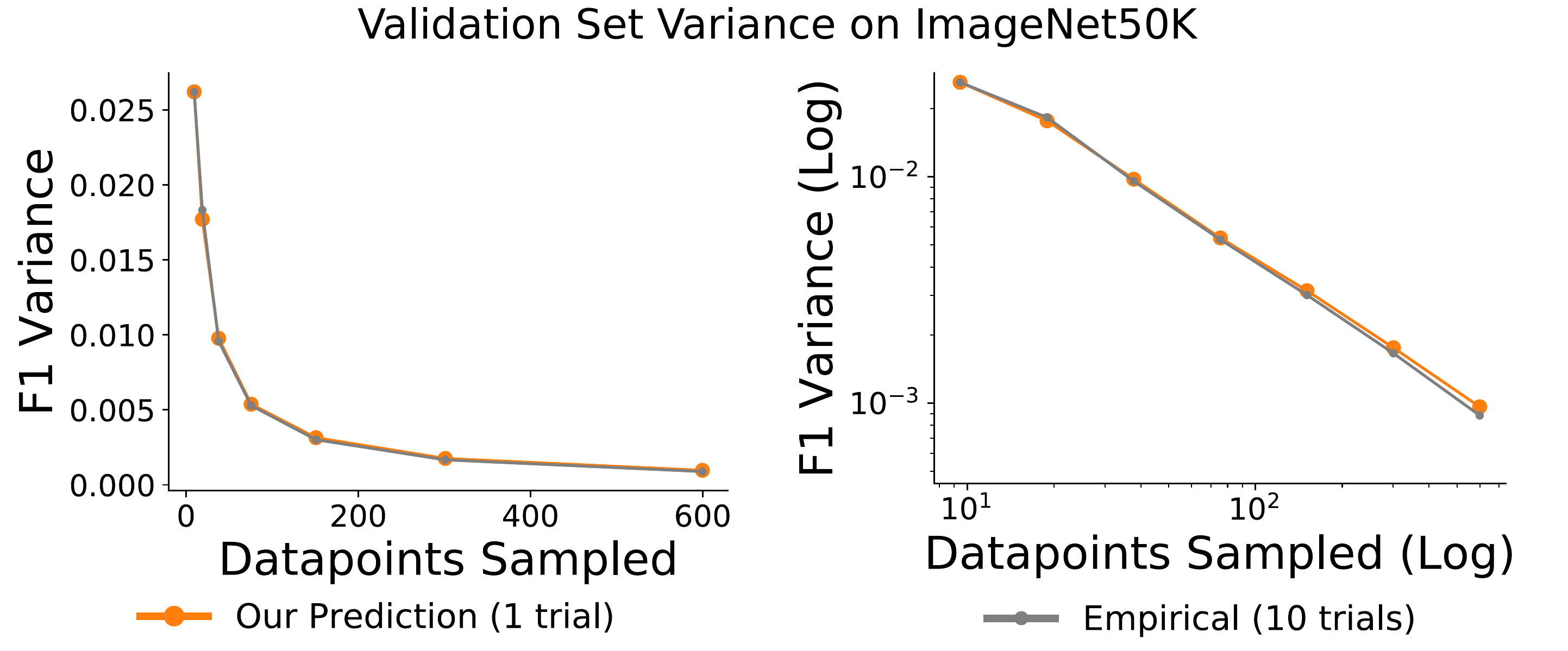}
    \end{subfigure}
    \end{center}
    \vspace{-1.5em}
    \caption{Our single-trial predictions of the variance of \ours\ (orange) closely mimic empirical estimates of its variance (gray) when evaluating SwAV. 
    We slightly underestimate the empirical variance in the very low-sample regimes, but quickly converge to the empirical estimates as sample size increases.  
    Reliable estimates of empirical variance make our estimates of F-score more practically useful. 
    For example, if aiming for an variance of $0.01$ in an F1 estimate, we know that the variance objective has been met after 40 samples.}
    \label{fig:plot2}
\end{figure}

We compare our estimate (Formula~\ref{eq:ours}) of the variance of \ours's F1 estimate to the method's empirical variance on \imagenetM\ (Fig.~\ref{fig:plot2}). 
Our method (orange line) closely approximates the empirical variance (gray line) of \ours\, even when computed from a small number of samples. 
The empirical variance is computed by performing 10 independent trials of \ours.

\vspace{-1.0em}
\subsubsection{Finite-Dataset Variance}
\vspace{-0.5em}

\begin{figure}[t]
    \begin{center}
      \begin{subfigure}{.96\linewidth}
        \includegraphics[width=\linewidth]{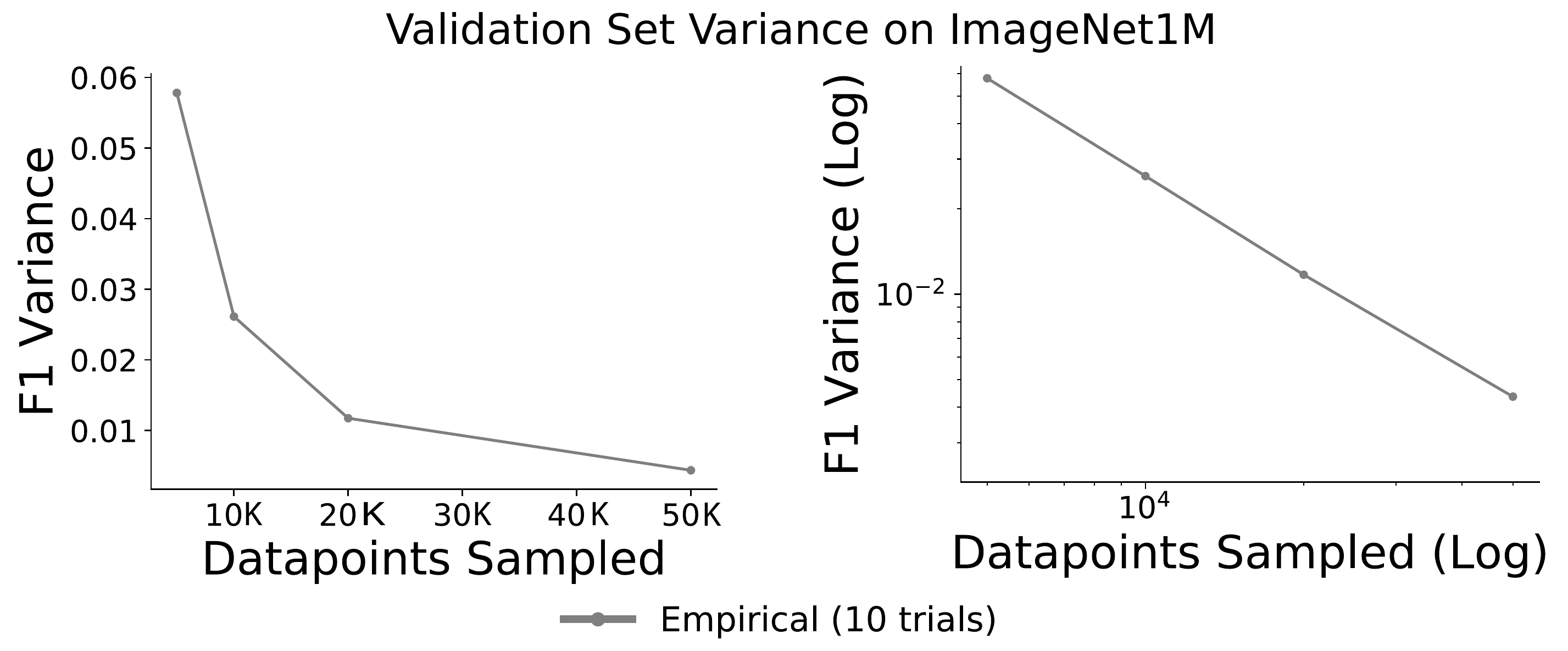}
      \end{subfigure}
    \end{center}
    \vspace{-1.5em}
    \caption{
        When validating SwAV on large, randomly sampled validation datasets, we observe significant variance in predicted F1 even when sampling up to 50,000 images.
        This suggests that even validation estimates calculated from traditional large datasets have notable uncertainty when validating binary classifiers trained for rare categories.
    }
    \label{fig:plot1}
\end{figure}

Computing F-score using a large, but finite, dataset will often yield a very good estimate of the F-score for the full test population. 
However, due to their finite size, even these estimates have variance, in particular when classifying rare categories. We can estimate this variance using Formula~\ref{eq:ours}.

Figure~\ref{fig:plot1} illustrates that notable variance in estimates remain present even when using large, randomly-sampled validation datasets.
To assess the accuracy of our variance estimate, for each of the 1000 ImageNet categories we randomly sample subsets (of up to 50,000 images) of the \imagenetM\ dataset, and we evaluate the F1 of SwAV on these validation sets.

The variance of the estimate of F1 on a 50,000-sample dataset (the size of the ImageNet validation dataset) is $0.003$, implying a standard devation of $\sqrt{0.003} \simeq 0.055$. 
\emph{Therefore, when evaluating binary classifiers trained on rare categories, many datasets commonly used to compute ground-truth estimates of model performance themselves contain notable uncertainty.} 

\vspace{-0.5em}
\subsection{Sharing Validation Sets Across Models}
\vspace{-0.5em}

\begin{figure}[t]
    \includegraphics[width=\linewidth]{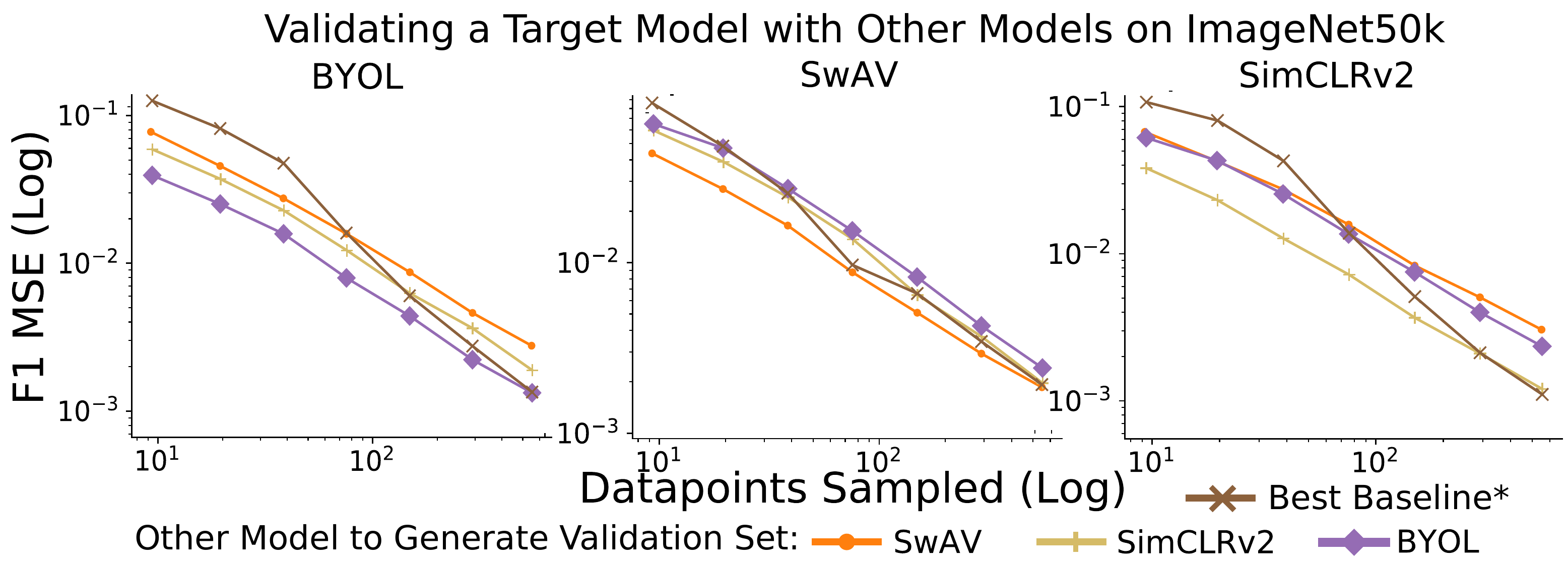}
    \vspace{-1.5em}
    \caption{Validation sets curated for each of SwAV, BYOL, and SimCLR can be used to efficiently estimate F1 score of the other two models on \imagenetK. 
    For a given sample size, the MSE of the \ours\ estimator tailored to a different model is at most double the MSE of the model-specific estimator.
    These estimators are competitive with, and sometimes outperform, the best model-specific baseline approaches in the low-sample regime. 
    *The brown lines reflect the best of the model-specific baseline methods from Figure \ref{fig:plot3}.
    }
    \label{fig:multimodel}
\end{figure}

The previous results demonstrate that validation efficiency can be significantly improved by curating a validation set for the \emph{specific model under evaluation}.  However, these labeled samples can be used to estimate F-scores for other models as well.
(Our estimator remains consistent when evaluating other models.)  
To understand how validation sets transfer to other models, for each of SwAV, SimCLRv2, BYOL we curate validation sets for estimating F1 performance of the model, then we use these validation sets to estimate the performance of all three models (Figure~\ref{fig:multimodel}).

For all three models, the F1 estimates have the lowest MSE when curated specifically for the desired model. However, F1 estimates generated from datasets curated for a given model are surprisingly effective for validating other models.
\ours\ datasets curated for different models obtain F1 estimates with MSE values competitive with the best baseline techniques that are actually tailored to the desired model (brown lines in Figure~\ref{fig:multimodel}), sometimes even outperforming the baseline techniques in the low-sample regime.
For instance, when validating SimCLRv2, an \ours\ dataset curated for SwAV performs better than the best baseline trained on SimCLRv2 itself when sampling less than 80 points. 
The surprising effectiveness of using a model-specific validation set to validate other models may be due in part to similarity between the models validated. 
Nevertheless, our results suggest that, for a similar family of models, actively curated validation sets can effectively estimate F-score across different models.

\section{Discussion}

We have presented a method for estimating the F-score of binary classification models on a low label budget, as well as a method for predicting the variance in this estimate.  Our approach constructs validation sets specifically for the target model to evaluate, but we demonstrate that validation sets constructed for one model can also be used to efficiently validate similar models.
This observation suggests that, for a given task, it might be possible to actively construct model-agnostic datasets that enable accurate validation with far fewer labeled samples than datasets typically used today. 

That said, for models that substantially differ from prior art, our experiments suggest that it may be worth constructing one-off model-{\em specific} validation sets. For example, safety-critical perception tasks that enable self-driving vehicles might require extremely precise estimates of validation performance of deployed models. Rather than validating models on pre-defined fixed test sets, our method may provide a framework for actively validating models on live on-fleet data streams.

Finally, our algorithm's accurate variance estimates can also be used to construct confidence intervals,
and also provide guidance on the sample budget required to obtain an estimate of F-score with an acceptable level of variance.
Future analysis analyzing the covariance of F-score estimates across models, combined with per-model variance estimates, might facilitate analysis of whether differences in model performance, potentially estimated on different datasets, are statistically significant or not.

\vspace{-1.5em}
\paragraph{Acknowledgments} This work is supported by the National Science Foundation (NSF) under III-1908727 and CCF-1937301, as well as the CMU Argo AI Center for Autonomous Vehicle Research. 
Vishnu Sarukkai is supported by the National Defense Science \& Engineering Graduate (NDSEG) Fellowship Program.

{\small
\bibliographystyle{ieee_fullname}
\bibliography{paper}
}

\raggedbottom

\pagebreak

\onecolumn
\begin{appendices}
\section{Training Details}
As mentioned in Section~4.1, we train the SwAV and BYOL models on the 1\% split of ImageNet as these pre-trained versions of these models are not provided by the authors. We also trained a SwAV model on the 10\% split of iNaturalist that we created to evaluate on \inat. We will release the 10\% split of iNaturalist with the rest of the code for the paper. Note that we attempt to train semi-supervised models that perform close to what is reported by the authors, but, for our experiments, it is more important that the models are representative of semi-supervised models trained for the target dataset (that is, they does not necessarily need to be state-of-the-art).

For SimCLRv2, we use the pretrained model provided by the authors trained on the 1\% split of ImageNet~\footnote{Model code: \url{https://github.com/google-research/simclr}. Training weights: \url{https://console.cloud.google.com/storage/browser/simclr-checkpoints/simclrv2/finetuned_1pct}}. 

For SwAV, we produced a semi-supervised model by fine-tuning the provided self-supervised weights on the 1\% split of ImageNet using the author's official code for semi-supervised learning~\footnote{\url{https://github.com/facebookresearch/swav\#evaluate-models-semi-supervised-learning-on-imagenet}}.
For evaluating on \inat, we modified the author's semi-supervised training code in the following manner: (1) we modified the data loader to read the 10\% iNaturalist training split, and (2) we changed the input image size to 299 from 224.

For BYOL, we modified the linear fine-tuning code provided by the authors~\footnote{\url{https://github.com/deepmind/deepmind-research/tree/master/byol\#linear-evaluation}} to (1) fine-tune the whole network by unfreezing the backbone (propagating gradient updates to all layers of the model during training), (2) train for 50 epochs instead of 80, and (3) to train on the 1\% split of ImageNet.

\section{Estimator Robustness to Classifier Thresholds}

Figure 9 from the main paper illustrates that labeled datasets sampled for a given model can be reused to compute F1 for different models that are performing the same classification task. 
This suggests that labeled datasets sampled by our algorithm can be reused effectively across model families. 
Here, we run a similar experiment on a model family constructed through varying the classifier threshold (the model score above which we label samples as positives). 

We construct binary classifiers by 1) taking the 1000-way outputs from the models described in Section 4.1 of the paper, 2) constructing probabilities per class by applying a softmax transformation to the logits, 3) converting the probabilities to positive and negative class probabilities through one-vs-all classification, then 4) calculating predicted labels by checking whether the positive class probability is greater than a threshold $t$. 

We use \oursmulti\ to estimate the F-Score for binary classifiers using a SwAV model on \imagenetK\ using the threshold $0.1$, producing a labeled dataset. 
We reuse this labeled dataset across a family of binary classifiers constructed using thresholds ranging from $0.02$ to $0.5$.

\begin{figure}[h]
    \centering
    \captionsetup{width=\linewidth}
    \includegraphics[width=0.5\linewidth]{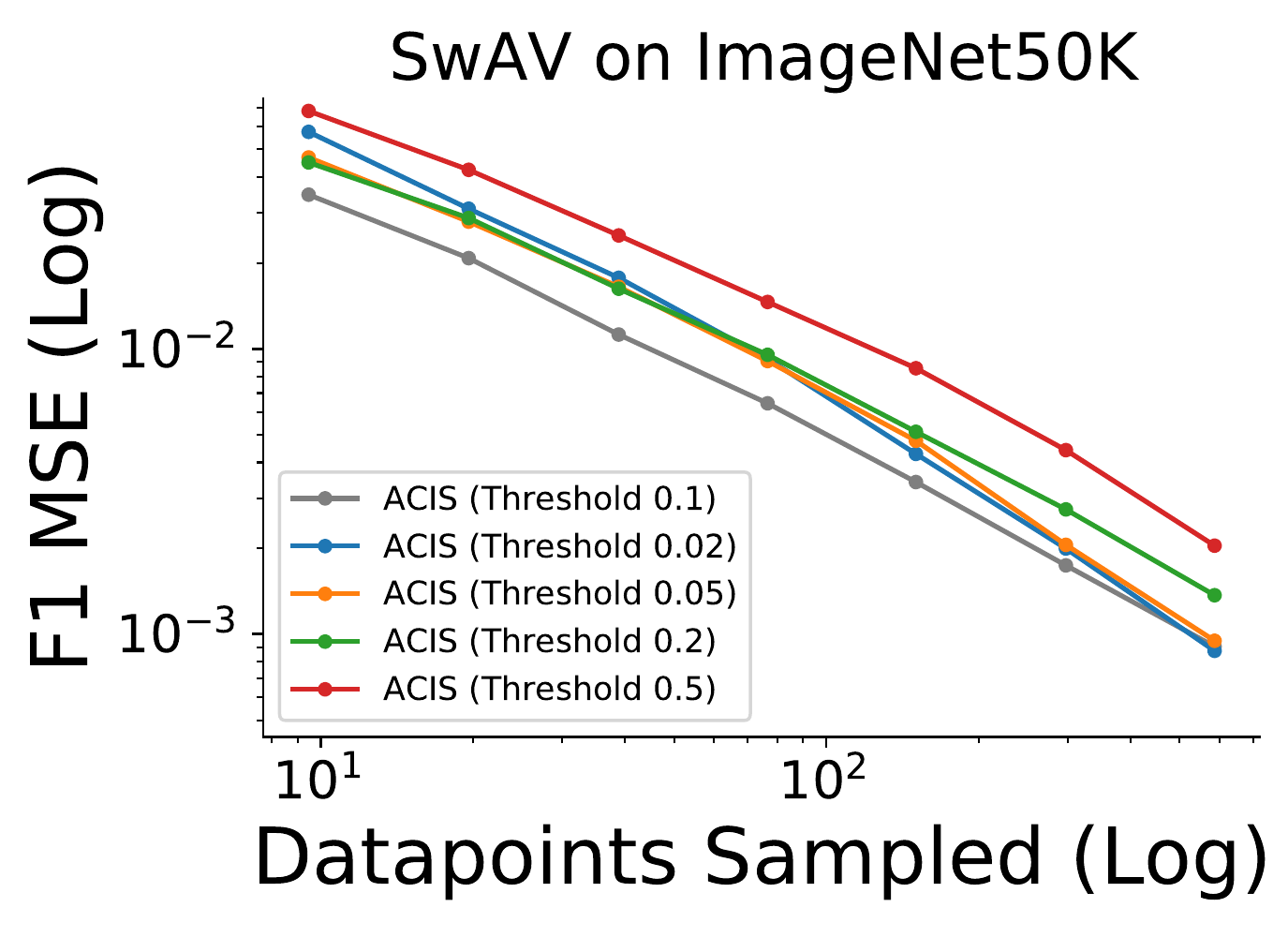}
    \caption{Validation sets curated for SwAV for a model with a given classifier threshold can be used to efficiently estimate F1 score of the model family generated by varying the classifier threshold. 
    The validation set curated for a binary classifier on \imagenetK\ using a threshold of $0.1$ accurately estimates the F1 of the models generated by using thresholds ranging from $0.02$ to $0.5$. The labeled datasets transfer slightly better to models with lower classifier thresholds than to models with higher classifier thresholds. 
    }
    \label{fig:thresholds}
\end{figure}

In Figure \ref{fig:thresholds}, the labeled datasets curated for a threshold of $0.1$ effectively estimate F1 across the range of thresholds tested, though performance is slightly worse for higher classifier thresholds. 
Along with the results from Section 4.4 of the paper, these results suggest that it is possible to curate an actively labeled validation set that can effectively estimate F-score across a family of models.
\section{Platt scaling vs isotonic regression}

\begin{figure}[h]
    \includegraphics[width=\linewidth]{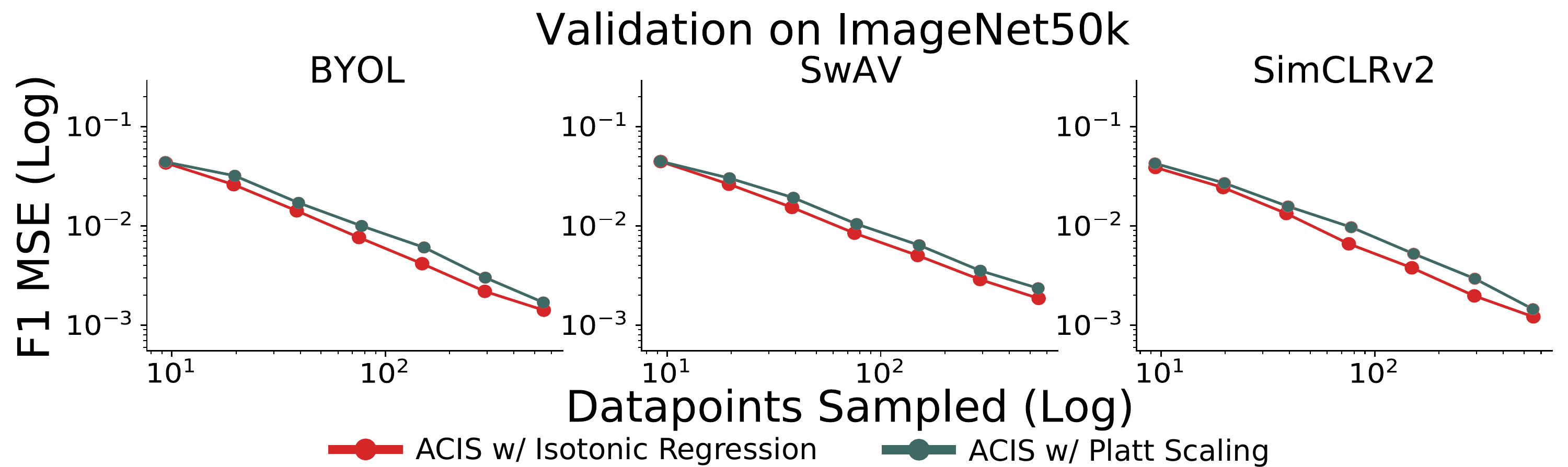}
    \caption{
        \ours\ performs better with isotonic regression than with Platt scaling. 
    }
    \label{fig:platt}
\end{figure}

Figure~\ref{fig:platt} shows the result of replacing isotonic regression in our method with Platt scaling.
As seen in this figure, Platt scaling performs worse than isotonic regression in our experiments.
Unlike Guo et al.~\cite{Guo2017NNCalibration}, who evaluated under fairly balanced categories and calibrated using thousands of labeled examples, our work calibrates binary classifiers with severe class imbalance using only tens of labeled examples.

\section{Variance Estimation}

\newcommand{\ep}{\epsilon}
\newcommand{\f}[2]{\dfrac{#1}{#2}}
\newcommand{\ff}[2]{\frac{#1}{#2}}
\newcommand{\R}{\mathbb{R}}
\newcommand{\ceil}[1]{\left\lceil{#1}\right\rceil}
\newcommand{\de}{\delta}
\newcommand{\T}{\theta}
\newcommand{\su}[2]{\mathlarger{\sum\limits_{#1}^{#2}}}
\newcommand{\pd}[2]{\mathlarger{\prod\limits_{#1}^{#2}}}
\newcommand{\E}{\mathbb{E}}
\newcommand{\X}{\mathcal{X}}
\newcommand{\Y}{\mathcal{Y}}
\newcommand{\D}{\mathcal{D}}
\newcommand{\tr}{\text{tr}}
\newcommand{\val}{\text{val}}
\newcommand{\re}{\text{ref}}
\renewcommand{\split}{\text{split}}
\newcommand{\lt}{\left(}
\newcommand{\rt}{\right)}
\newcommand{\Lt}{\left[}
\newcommand{\Rt}{\right]}
\newcommand{\A}{\alpha}
\newcommand{\N}{\mathcal{N}}
\newcommand{\n}{\eta}
\newcommand{\B}{\beta}
\newcommand{\M}{\lambda}
\newcommand{\arr}[1]{\lt\begin{array}{#1}}
\newcommand{\I}[2]{\mathlarger{\int_{#1}^{#2}}}
\newcommand{\G}{\nabla}
\newcommand{\C}{\mathbb{C}}
\newcommand{\1}{\mathbbm{1}}
\newcommand{\ra}{\rightarrow}
\newcommand{\Lim}[1]{\lim\limits_{#1}}
\newcommand{\Sup}[1]{\sup\limits_{#1}}
\newcommand{\floor}[1]{\lfloor #1\rfloor}
\newcommand{\w}{\omega}
\providecommand{\norm}[1]{\lVert#1\rVert}
\renewcommand{\P}{\mathcal{P}}
\newcommand{\te}{\triangleq}
\renewcommand{\d}[1]{\,\mathrm{d}#1}
\renewcommand{\l}{\ell}

\subsection{Averaging Across Iterations}

\begin{figure}[h]
    \centering
    \captionsetup{width=\linewidth}
    \includegraphics[width=0.5\linewidth]{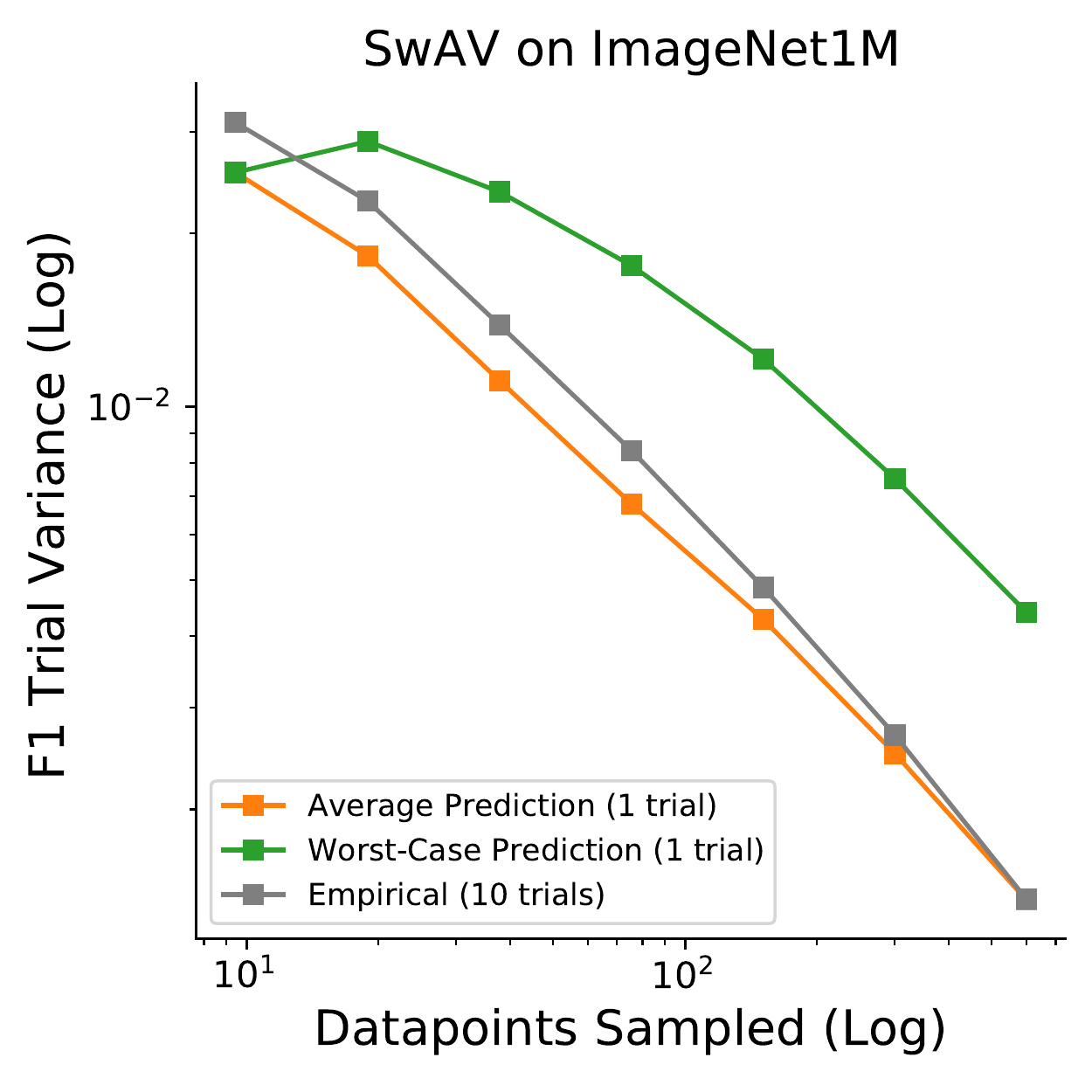}
    \caption{When estimating the F-score of a SwAV model trained on \imagenetM, a weighted average of variance estimates across iterations for \oursmulti\ produces a variance estimate which is very similar to the observed empirical variance across trials. On the other hand, assuming a covariance of $1$ across trials and estimating a worst-case variance for \oursmulti\ significantly overestimates the observed empirical variance.}
    \label{fig:variance_ablation}
\end{figure}

In Section 3.4, we note that when combining the estimated $\hat{G}_j$'s, we estimate the variance of the weighted average $\hat{G}$ by taking a weighted average of the sampling variances of each iteration. 
While this assumes that there is no covariance between the $\hat{G}_j$'s, we find that this is a reasonable assumption in practice, and yields better estimates of the variance of $\hat{G}$ than the worst-case estimator (which assumes a covariance of 1).
As seen in Figure \ref{fig:variance_ablation}, the worst-case estimator significantly overestimates the empirical variance, whereas taking a weighted average of the sampling variances across iterations yields an accurate estimator of the empirical variance. 

\subsection{Derivation of Estimator}
\newcommand{\Var}{\text{Var}}
\newcommand{\Cov}{\text{Cov}}
\renewcommand{\l}{\ell}

In this section, we motivate the definition of our estimator
$$S^2_{n,q} = \frac {C^{-1} * \sum_{j=1}^n \left( w(x_j,y_j,M(x_j))^2 \left( \ell(M(x_j),y_j) - \hat{G}_{n,q}\right)^2  \right)} {\frac{1}{n}\left(\sum_{j=1}^n w(x_j,y_j,M(x_j))\right)^2}$$
of the sampling variance introduced in Section~3.4.
The derivation below largely follows that of Lemma 2 of Sawade et al.~\cite{Sawade2010ActiveFMeasures}, correcting an error in their variance derivation.
Let $q(x,y)$ be any distribution such that $q(x, y)$ can be factorized as $q_x(x) p(y|x)$ for some $q_x(x)$.
As in the main paper, we use the notation $q(x)$ to denote $q_x(x)$ and similarly use $p(x)$ to denote
the marginal distribution of $x$ under $p$.

Define
$\hat{G}_{n,q} =  \f{\sum_{j=1}^n \ell(x_j,y_j,\hat{y}_j)w(x_j,y_j,\hat{y}_j)}
{\sum_{j=1}^n w(x_j,y_j,\hat{y}_j)}$,
where $\{(x_j,y_j)\}_{j=1}^n$ are drawn IID according to $q(x,y)$ and $\hat{y}_j = M(x_j)$ [where $M$ is the model].
We first claim that the asymptotic variance of $\hat{G}_{n,q}$ is
{\small
$$\f{\I{}{} \I{}{} w(x,y,M(x))^2 \lt \l(M(x), y) - G\rt^2 q(x,y) \, dy \, dx}{\lt \mathlarger{\iint} w(x, y,M(x)) q(x,y) \,dy\, dx \rt^2}$$
}

\noindent\emph{Proof}: Define
\begin{align*}
v_j &= v(x_j,y_j, \hat{y}_j) = v(x_j,y_j,M(x_j)), \\
w_j &= \tfrac{p(x_j)}{q(x_j)} \cdot v_j, \\
\l_j &= \l(\hat{y}_j, y_j),\\
\hat{G}^0_{n,q} &= \sum_{j=1}^n \ell_j w_j,\\
W_n &= \sum_{j=1}^n w_j~.
\end{align*}
Thus, $\E_q[\hat{G}^0_{n,q}] = \sum_{j=1}^n \E_q[ \ell_j w_j] = n \E_q[ \ell_j w_j]$.

\noindent $\E_q[ \ell_j w_j] = \E_q \Lt \ff{p(x_j)}{q(x_j)} v_j \l_j \Rt = \E_p[v_j \l_j]$.
Defining $G = \f{\E_p[v_j \ell_j]}{\E_p[v_j]}$, we thus have $\E_q[w_j \ell_j] = G\, \E_p[v_j]$. Also, $\E_q[W_n] = n \E_q[w_j] = n \E_p[v_j]$.
By the Central Limit Theorem, we thus have \begin{align*}
\sqrt{n} \lt \ff{1}{n} \hat{G}^0_{n,q} - G\E_p[v_j] \rt &\overset{D}{\ra} \N(0, \Var_q(w_j\ell_j)) \\\sqrt{n} \lt \ff{1}{n} W_n- \E_p[v_j] \rt &\overset{D}{\ra} \N(0, \Var_q(w_j)) 
\end{align*}

\noindent Define $f(x,y) = \ff{x}{y}$, so $\G f(x, y) = \lt \ff{1}{y}, \, -\ff{x}{y^2}\rt^T$. So, $f(G\E_p[v_j], \E_p[v_j]) = G$ and $\G f(G\E_p[v_j], \E_p[v_j]) = \lt \ff{1}{\E_p[v_j]}, -\ff{G\E_p[v_j]}{\E_p[v_j]^2}\rt^T \\=
\ff{1}{\E_p[v_j]} (1, -G)^T$.
Applying the Delta Method to the random vector $\ff{1}{n} (\hat{G}^0_{n,q}, \, W_n)^T$ and the function $f$, we thus have
\begin{align*}
\sqrt{n} \lt \hat{G}^0_{n,q} / W_n - G \rt \overset{D}{\ra} 
\N\lt 0, \tfrac{1}{\E_p[v_j]^2} (1, -G) \cdot \Sigma \cdot (1, -G)^T \rt,
\end{align*}
where $\Sigma = \lt \begin{array}{cc} \Var_q(w_j\ell_j) & \Cov_q(w_j\l_j, w_j) \\ \Cov_q(w_j\l_j, w_j) & \Var_q(w_j) \end{array}\rt$.

\noindent Now, $(1, -G) \cdot \Sigma \cdot (1, -G)^T =$
{\small
\begin{align*}
 & (1, -G) \cdot \lt\begin{array}{c}  \Var_q(w_j\ell_j )  - G\Cov_q(w_j\l_j, w_j) \\ \Cov_q(w_j\l_j, w_j) - G\Var_q(w_j ) \end{array}\rt = \\
& \Var_q(w_j\ell_j ) - 2G\Cov_q(w_j\l_j, w_j)  + G^2 \Var_q(w_j ) = \\
& ( \E_q[w_j^2\ell_j^2 ] - \E_q[w_j \ell_j ]^2 ) - 2G ( \E_q[w_j^2 \l_j] - \E_q[w_j \ell_j ] \E_q[w_j ] ) \\
&+ G^2 ( \E_q[w_j^2 ] - \E_q[w_j]^2 ) = \\
& ( \E_q[w_j^2 \ell_j^2 ] - 2 G \E_q[w_j^2 \ell_j ] + G^2 \E_q[w_j^2 ] ) - \\
&( \E_q[w_j \l_j ]^2 - 2G \E_q[w_j \l_j ]\E_p[w_j ] + G^2 \E_q[w_j ]^2) = \\
& \E_q[w_j^2 (\ell_j - G)^2]  - \lt \E_q[w_j \l_j ] - G \E_q[w_j ] \rt^2.
\end{align*}
}

\noindent Recall that $\E_q[ w_j] = \E_p[v_j]$ and $\E_q[ \l_j w_j] = G \E_p[v_j]$, so $\E_q[w_j \l_j ] - G \E_q[w_j ]  = G \E_p[v_j] - G \E_p[v_j] = 0$.
So, we finally obtain that the asymptotic variance is
$\tfrac{1}{\E_p[v_j]^2} (1, -G) \cdot \Sigma \cdot (1, -G)^T = \tfrac{1}{\E_p[v_j]^2}\E_q[ w_j^2 (\ell_j - G)^2] =$
$$\dfrac{\mathlarger{\iint} w(x,y,M(x))^2 (\l(M(x), y)-G)^2 \cdot q(x,y) \,dy\, dx}{\lt \mathlarger{\iint} p(x,y) v(x, y, M(x)) \,dy\, dx\rt^2} =
\dfrac{\mathlarger{\iint} w(x,y,M(x))^2 (\l(M(x), y)-G)^2 \cdot q(x,y) \,dy\, dx}{\lt \mathlarger{\iint} w(x, y, M(x))q(x,y) \,dy\, dx\rt^2}~,$$
since $\ff{p(x,y)}{q(x,y)} = \ff{p(x)}{q(x)}$.\hfill \ensuremath{\Box}
\\\\
\noindent Define
\begin{equation}
  \tilde{S}^2_{n,q} := \frac { \ff{1}{n}\sum_{j=1}^n  w(x_j,y_j,M(x_j))^2 \left( \ell(M(x_j),y_j) - \hat{G}_{n,q}\right)^2} {\left(\frac{1}{n}\sum_{j=1}^n w(x_j,y_j,M(x_j))\right)^2}
\end{equation}
By Slutsky's theorem and the preceding derivation, $\tilde{S}^2_{n,q}$ is a consistent estimator of the asymptotic variance of $\hat{G}_{n,q}$.

Now, let $C = 1 - \dfrac{\sum_{j=1}^n w(x_j,y_j,M(x_j))^2}{(\sum_{j=1}^n w(x_j,y_j,M(x_j)))^2}$. By Slutsky's theorem, $\dfrac{\ff{1}{n}\sum_{j=1}^n w(x_j,y_j,M(x_j))^2}{n\lt \ff{1}{n}\sum_{j=1}^n w(x_j,y_j,M(x_j))\rt^2}$
converges in probability to 0 as $n \ra \infty$, and hence $C$ converges in probability to 1. Thus, $S^2_{n,q} = \f{\tilde{S}^2_{n,q}}{C}$ is also a consistent estimator of the asymptotic variance of $\hat{G}_{n,q}$,
by applying Slutsky's theorem one final time.

The factor $C$ can be interpreted analogously to the Bessel correction for sample variance. Indeed, if the $w_j$'s are all equal to 1 (so $\hat{G}_{n,q}$ is simply the mean of the $\l_j$'s),
we have $C = 1 - \f{n}{n^2} = \f{n-1}{n}$ and $\tilde{S}_{n,q}^2 = \f{\sum_{j=1}^n (\l_j - \hat{G}_{n,q})^2}{n}$, so $S_{n,q}^2 = \ff{1}{n-1} \sum_{j=1}^n (\l_j -\hat{G}_{n,q})^2$,
which is the same as the standard sample variance calculation using Bessel's correction.
\end{appendices}

\end{document}